\documentclass[10pt,twocolumn,letterpaper]{article}

\usepackage{cvpr}
\usepackage{times}
\usepackage{epsfig}
\usepackage{graphicx}
\usepackage{amsmath}
\usepackage{amssymb}

\usepackage{subcaption}

\usepackage[breaklinks=true,bookmarks=false]{hyperref}

\cvprfinalcopy 

\def\cvprPaperID{26} 

\begin{document}

\title{Exemplar Guided Face Image Super-Resolution without Facial Landmarks}

\author{Berk Dogan, Shuhang Gu, Radu Timofte\\
Computer Vision Lab, D-ITET, ETH Zurich\\
{\tt\small doganb@student.ethz.ch, shgu@ee.ethz.ch, radu.timofte@vision.ee.ethz.ch}
}

\maketitle

\begin{abstract}
Nowadays, due to the ubiquitous visual media there are vast amounts of already available high-resolution (HR) face images. Therefore, for super-resolving a given very low-resolution (LR) face image of a person it is very likely to find another HR face image of the same person which can be used to guide the process. In this paper, we propose a convolutional neural network (CNN)-based solution, namely GWAInet, which applies super-resolution (SR) by a factor $8\times$ on face images guided by another unconstrained HR face image of the same person with possible differences in age, expression, pose or size. GWAInet is trained in an adversarial generative manner to produce the desired high quality perceptual image results. The utilization of the HR guiding image is realized via the use of a warper subnetwork that aligns its contents to the input image and the use of a feature fusion chain for the extracted features from the warped guiding image and the input image. In training, the identity loss further helps in preserving the identity related features by minimizing the distance between the embedding vectors of SR and HR ground truth images. Contrary to the current state-of-the-art in face super-resolution, our method does not require facial landmark points for its training, which helps its robustness and allows it to produce fine details also for the surrounding face region in a uniform manner. Our method GWAInet produces photo-realistic images in upscaling factor $8\times$ and outperforms state-of-the-art in quantitative terms and perceptual quality.
\end{abstract}


\section{Introduction}
Face image super-resolution or face hallucination aims at reconstructing details / high-frequencies in low-resolution (LR) face images. This is an important problem due to the increasing need for high-resolution (HR) face images for different applications such as security, surveillance or other application that involves face recognition.


\begin{figure}[th]
\centering
\setlength{\tabcolsep}{1pt}
\resizebox{\linewidth}{!}
{
\begin{tabular}{cccc}
\includegraphics[width=.4\linewidth]{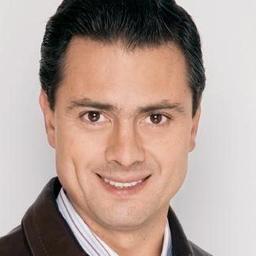}&
\includegraphics[width=.05\linewidth]{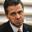}&
\includegraphics[width=.4\linewidth]{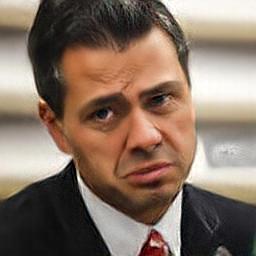}&
\includegraphics[width=.4\linewidth]{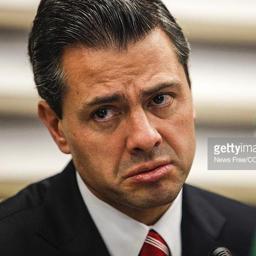}\\
guiding face& input &our result & ground truth\\
\end{tabular}
}
\caption{Exemplar guided face image super-resolution result ($8\times$) of our proposed GWAInet approach.}
\label{fig:comp_intro}
\end{figure}

Due to the increasing interest in visual media and the development of the social media, it is very likely that given a LR face image of a person, we can find another HR face image of the same person possibly taken at a different time in different conditions. This guiding face could be used in the super-resolution (SR) process to guide the hallucination of high frequencies/details, which might increase the quality of the HR result and help to preserve the identity related features. Fig.~\ref{fig:comp_intro} shows such a case and our result.

The current state-of-the-art face image super-resolution approach~\cite{Li_2018_ECCV} proposed the use of a guiding image together with a facial landmark detector, where an additional loss term is optimized such that the facial landmarks of the warped guiding image and those of the ground truth image are close to each other. However, this approach seems to produce fine details for the face region in a non-uniform and unpredictable manner, resulting in SR images that look only partially sharp.

Although the recently proposed CNN-based SR solutions~\cite{NTIRE2017, NTIRE2018} provide state-of-the-art quantitative results in terms of peak signal-to-noise ratio (PSNR) when they optimize for reconstruction losses such as L1 or L2 in image space, the results are smooth without the fine details required for a good perceptual quality. This problem is more visible with the increase of the upscaling factor~\cite{2016arXiv160308155J, blau20182018}. On top of that, the PSNR measure is unable to capture perceptually important differences between two images as it relies on the differences between pixel-level values at the same position~\cite{Wang04imagequality, 1292216, 6146669}. One way to introduce perceptually important features into the SR image is to use generative adversarial networks (GANs)~\cite{2014arXiv1406.2661G,8099502,blau20182018}. These networks help to create realistic SR images that look like HR images, which are naturally sharper and contain fine details.

In this paper, we introduce a novel CNN architecture capable of generating high quality HR face images with an upscaling factor $8\times$. During the SR process, the network utilizes the LR face image and the extra information provided by another HR face image of the same person while making the necessary processing through a warping subnetwork on this guiding HR image. 
By addressing the possible differences in contents (\eg expression, pose, size) between two images the warper facilitates the extraction and integration of information from the guiding image. Contrary to the current state-of-the-art approach~\cite{Li_2018_ECCV}, our method does not require facial landmarks during training. This makes the network to learn and process the whole face region in a uniform manner and adds robustness. We add also an identity loss to further help in preserving the identity related features by minimizing the distance between the embedding vectors of HR result and ground truth images. Utilization of the guiding image expresses itself qualitatively as an improvement in visual content quality by correcting the inaccurate facial details. Finally, the adversarial loss that is incorporated via a GAN setting, will introduce fine details to the SR image and produce face images that are hardly distinguishable from real HR face images.

\section{Related Work}
\textbf{Convolutional Neural Networks (CNNs) and Image Super-Resolution (SR).} CNNs have emerged as a successful method in many computer vision applications~\cite{Bengio:2009:LDA:1658423.1658424, Krizhevsky:2012:ICD:2999134.2999257, 43022}. Deep learning with CNNs has also become very widely used in image SR~\cite{Kim2016AccurateIS, Dong:2016:ISU:2914182.2914303, 8099502, 2017arXiv170702921L}. CNNs have outperformed previous works~\cite{yang2010,zeyde2010single,Timofte_2013_ICCV,timofte2014a+,Agustsson_2017_CVPR_Workshops} both quantitatively and qualitatively. These networks, on the other hand, are mainly used for SR of single or multiple LR images and do not utilize a guiding HR image for the given LR image. In our work, we use the network given in~\cite{2017arXiv170702921L} as the main structural element of our subnetworks and specifically work on face images while utilizing the additional information provided by another HR face image of the same person.  

\textbf{Spatial Transformer Networks.} Spatial transformer networks are modules that can be incorporated into an existing network and trained in an end-to-end fashion without any modification to the learning scheme or the loss function ~\cite{2015arXiv150602025J}. They increase the spatial invariance of the network and provide invariance for large transformations~\cite{2015arXiv150602025J}. They are used to spatially transform input feature maps and consist of a localisation network and a sampler. In our work, we use the ideas from spatial transformer networks to create a flow field, which is then used in combination with a bilinear sampler to warp the guiding image, thus making the guiding image aligned with the contents of the input image.

\textbf{Face Hallucination.} CNNs have also shown great success in the field of face hallucination, where we apply super-resolution on face images~\cite{Yu2016UltraResolvingFI, AAAI1714340, Zhou:2015:LFH:2888116.2888253, Li_2018_ECCV, 10.1007/978-3-319-46454-1_37, 8237298, 2017arXiv170803132C, Zhang_2018_ECCV, Yu2017HallucinatingVL}. \cite{Yu2016UltraResolvingFI} applies face hallucination on tiny $16\times16$ faces. In \cite{AAAI1714340, Yu2017HallucinatingVL}, the authors again work on tiny $16\times16$ images but use spatial transformer networks~ \cite{2015arXiv150602025J} in their generator architecture to alleviate the effects of misalignment of input images. Zhou~\etal~\cite{Zhou:2015:LFH:2888116.2888253} fuse two channels of information, namely extracted facial features and the LR input image, in order to overcome problems related with appearance variations and misalignment. However, they use resource intensive fully-connected layers in upscaling process and follow a simple fusion operation by just summing the upscaled LR input image via bicubic interpolation with the HR image created from the facial features. In our approach, on the other hand, we cope with the effects of appearance variations through the use of spatial transformer networks. However, contrary to~\cite{AAAI1714340, Yu2017HallucinatingVL}, we introduce the spatial transformer network as a subnetwork that is only applied to the input image rather than intermediate feature maps and contrary to~\cite{Zhou:2015:LFH:2888116.2888253}, we use resource efficient convolutional layers in upscaling process and follow a complex feature fusion for the information coming from two channels. Most importantly, these works do not incorporate the use of an additional HR image of the same person. In a very recent work~\cite{Li_2018_ECCV}, Li~\etal use a guiding image and a warper subnetwork to cope with appearance variations between the LR input and the HR guiding image. However, they apply direct concatenation of warped guiding image and upscaled LR image at the input of the generator, which is different than our feature extraction and fusion based approach through the use of secondary feature extractor subnetwork, which is called GFEnet, for the warped guiding image. They also use landmark loss and total variation loss for their warping subnetwork in the joint training phase, whereas we do not incorporate these losses in our overall objective, thus our network does not require facial landmarks during training. Another difference is that they use conditional adversarial networks~\cite{2016arXiv161107004I} for generating the adversarial loss, whereas we use a Wasserstein generative adversarial network with gradient penalty (WGAN-GP) \cite{2017arXiv170400028G}. \cite{Li_2018_ECCV} is the only recent paper known to us that uses an additional guiding image in face hallucination. As in our approach, Zhang~\etal~\cite{Zhang_2018_ECCV} also use an identity loss in face hallucination problem, which is calculated between the SR image and the ground truth image.

\begin{figure*}[th]
\centering
\includegraphics[width=1.0\linewidth]{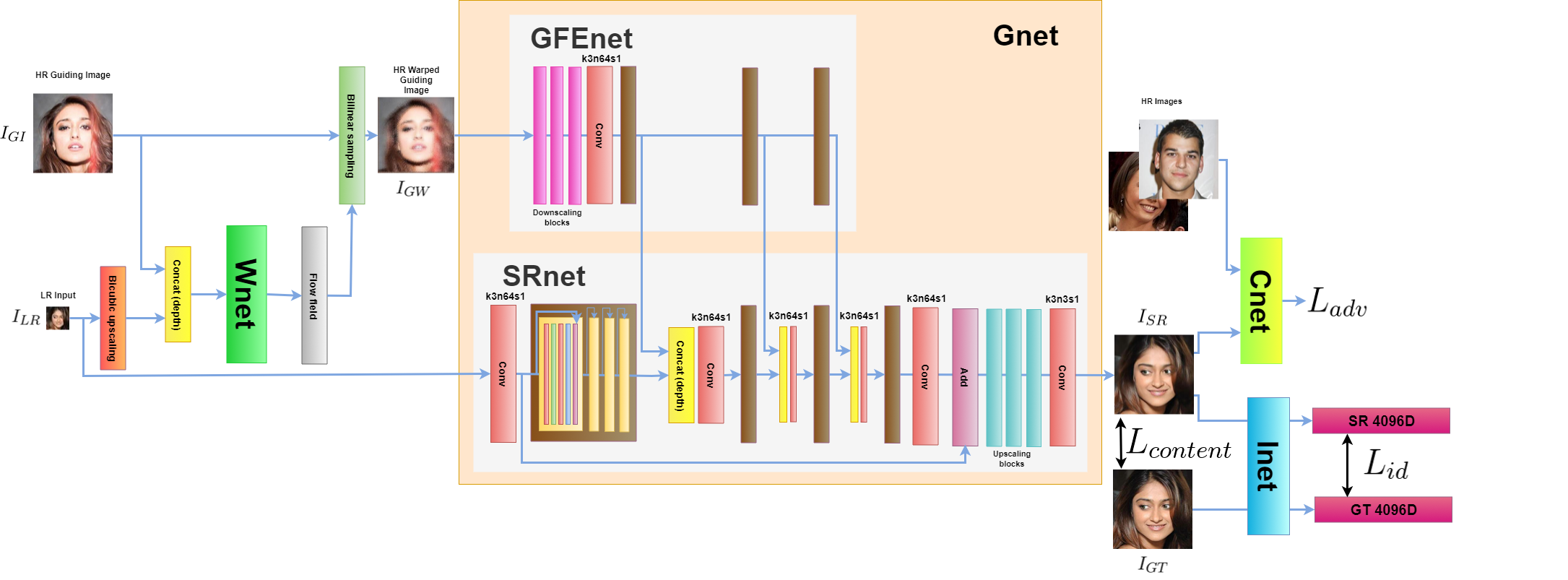}
\vspace{-0.2cm}
\caption{Proposed GWAInet and its Warper (Wnet), Generator (Gnet), Critic (Cnet) and Identity Encoder (Inet) subnetworks.}
\label{fig:full_model_identity}
\vspace{-0.2cm}
\end{figure*}

\textbf{Generative Adversarial Networks (GANs).} Although the SR methods using CNN architectures provide state-of-the-art quantitative results in PSNR terms when optimized for reconstruction losses such as L1 or L2 in image space, they produce overly-smooth visuals and lack the ability to produce images with fine details. The PSNR metric does not correlate well with the human perception of image quality~\cite{6622760}. This is due to the fact that the reconstruction loss is calculated in image space and the optimum solution is the average of all possible solutions~\cite{2016arXiv160202644D, 8099502, 2015arXiv151105666B}. GANs~\cite{2014arXiv1406.2661G} have become successful in creating realistically looking images thanks to their adversarial loss. As a result of this, many methods make use of GANs. \cite{2016arXiv160202644D} uses both a loss in feature space and an adversarial loss, in addition to the reconstruction loss in image space to generate sharp and natural looking images. Besides adversarial loss and reconstruction loss in image space, \cite{2015arXiv151105440M} uses an additional image gradient difference loss between the input and the output that sharpens the image prediction to predict future images from a video sequence. In \cite{Yu2016UltraResolvingFI, AAAI1714340}, they use adversarial loss and pixel loss to super-resolve tiny face images such that the resulting SR images have high frequency components. In \cite{8099502}, they use adversarial loss and feature loss for VGG-19 \cite{Simonyan14c} network to produce sharp and photo-realistic SR images. In \cite{Li_2018_ECCV}, they also include adversarial loss in their total loss to improve the output visual quality of face restoration tasks from degraded observations. In our work, we use adversarial loss together with the L1 reconstruction loss in image space. Reconstruction loss drives the networks to match the contents of the output SR image with the contents of the input LR image. The adversarial loss, on the other hand, tries to ensure that the SR image contains high frequency features that make it photo-realistic. As a result of this loss combination, we get an SR image that agrees with the input LR image in terms of facial feature location and the coarse specifications for these facial features but also agrees with the specifications imposed by the distribution of the HR face images. We specifically use WGAN-GP, which optimizes for a different metric than the traditional GANs and is found to be more stable and easier to train~\cite{2017arXiv170400028G}.

\section{Proposed Method}

Our proposed GWAInet solution (Guidance, Warper, Adversarial loss, Identity loss network) produces a SR image $I_{SR}$ from a LR input image $I_{LR}$ and a HR guiding image $I_{GI}$. $I_{LR}$ is obtained from a ground truth high-resolution image $I_{GT}$ by downscaling with bicubic interpolation in scale $8\times$. $I_{GI}$ is another HR face image of the person, to whom the tuple $(I_{LR},I_{GT})$ belongs to. We also denote the image that is obtained by warping $I_{GI}$ as $I_{GW}$. 

In the following, we provide detailed information about our GWAInet method. First, we briefly describe the WGAN-GP and then we present the network architecture of our model. Finally, we describe the loss functions used in guiding the optimization process. 

\subsection{WGAN-GP~\cite{2017arXiv170400028G}}

We use a generative adversarial network (GAN) approach~\cite{2014arXiv1406.2661G} to generate perceptually good and sharp images. Specifically, we use a Wasserstein GAN with gradient penalty (WGAN-GP)~\cite{2017arXiv170400028G}. With the help of this new architecture, we are aiming to make the SR images from our proposed network GWAInet as indistinguishable as possible from the HR images in our dataset. This is possible due to the structure of WGAN-GP and its training objective.

GANs consist of a generator subnetwork $G$ and a discriminator subnetwork $D$, where the aim of $G$ is to create samples that are as close as possible to the real data samples and the aim of the $D$ is to classify these fake samples from the real ones. If we denote the real data distribution by $p(x)$ and generated data distribution by $q(y)$, then objective can be formulated as~\cite{2014arXiv1406.2661G}:
\begin{equation}
\label{eq:gan_equation}
\begin{split}
min_{G}\; max_{D}\mathbf{E}_{x'\sim p(x)}\left [ logD(x')\right ]+\\\mathbf{E}_{y'\sim q(y)}\left [ 1-logD(y')\right ]
\end{split}
\end{equation}

In the traditional GAN setting, for an optimal discriminator, we are trying to minimize the JS-divergence between the real and generated data distributions~\cite{2014arXiv1406.2661G}. With this approach, training the model is a difficult process. This is due to the fact that in many practical problems, the real and the generated data distribution are disjoint in some low dimensional manifold in a high dimensional space, which makes it easier to find a perfect discriminator~\cite{2017arXiv170104862A}. When the discriminator becomes perfect, the gradient coming from the JS-divergence vanishes~\cite{2014arXiv1406.2661G}. There is an alternative method to avoid vanishing gradients by maximizing $\mathbf{E}_{y'\sim q(y)}\left [-logD(y')\right ]$ for $G$, however it is shown that this method causes unstable updates ~\cite{2017arXiv170104862A}. 

Wasserstein generative adversarial network (WGAN) introduces a new loss function for GAN training, which depends on the Earth-Mover distance~\cite{2017arXiv170107875A} formulated as:
\begin{equation} 
\label{eq:em_distance}
W(p,q)=inf_{\gamma \in \prod (p,q)}\mathbf{E}_{(x',y')\sim \gamma(x,y) }\left [ \left \| x'-y' \right \| \right ]
\end{equation}
In equation \ref{eq:em_distance}, $\prod$ denotes the set of all joint distributions $\gamma (x,y)$, where $\sum_{y}\gamma (x,y)=p(x)$ and $\sum_{x}\gamma (x,y)$ $=q(y)$. $\gamma (x,y)$ can be seen as the amount of earth that should be transported from $x$ to $y$ to transform $p$ into $q$. WGAN-GP is the improved version of WGAN, with an addition of gradient penalty term in the cost function instead of weight clipping procedure ~\cite{2017arXiv170400028G}. 

\subsection{Network Architecture}
The complete network architecture of the proposed solution is illustrated in Figure~\ref{fig:full_model_identity}. The complete model, called GWAInet, consists of four network components, namely Gnet, Wnet, Cnet and Inet.

\textbf{Warper (Wnet).} 
Wnet's aim is to produce the flow field required to warp the guiding image such that it is well aligned with the contents of the input LR image, removing any difference in pose or size of the faces in both images. This warping procedure allows the extra information provided by $I_{GI}$ to be better utilized. Wnet is essentially the localisation network for a spatial transformer network~\cite{2015arXiv150602025J}. It produces the transformation parameters that is fed into the bilinear sampler along with $I_{GI}$. Before the first layer of Wnet, upscaling via bicubic interpolation is applied to $I_{LR}$, which scales the spatial dimensions by 8, producing the image $I_{LRU}$. After that, $I_{GI}$ and $I_{LRU}$ are concatenated along the depth axis. The resulting tensor forms the input of Wnet. Wnet outputs a 3D flow field with 2 depth channels. At each pixel location, the first value determines the sampling motion horizontally and the second value determines the sampling motion vertically. It should be noted that flow field values are not scaled into a specific range, meaning that no constraints are applied at the output. The flow field and $I_{GI}$ are used in the bilinear sampling module to produce the warped guiding image $I_{GW}$. Let us denote the flow field as $\Omega\in \mathbb{R}^{h_{HR}\times w_{HR}\times 2}$ and denote the pixel location grid for $I_{GW}$ as $\delta\in \mathbb{R}^{h_{HR}\times w_{HR}\times 2}$, where $\delta (i,j,0) = \frac{2\times i}{h_{HR}-1}-1$ and $\delta (i,j,1) = \frac{2\times j}{w_{HR}-1}-1$ $\forall i\in \left \{ 0,1,...,h_{HR}-1 \right \}$ and $\forall j\in \left \{ 0,1,...,w_{HR}-1 \right \}$. Note that the grid values are in the range $\left [ -1,+1 \right ]$ instead of $\left [ 0, h_{HR}-1 \right ]$ for $\delta (i,j,0)$ and $\left [ 0, w_{HR}-1 \right ]$ for $\delta (i,j,1)$. In other words, in our setting, we assume that the top left corner of the image has coordinates (-1,-1) and the bottom right corner of the image has coordinates (+1,+1). Using $\Omega$ and $\delta$, the sampling grid $\rho\in \mathbb{R}^{h_{HR}\times w_{HR}\times 2}$ can be calculated as \cite{2015arXiv150602025J}: 
\begin{equation}
\label{eq:sampling_grid}
\begin{split}
\rho=\Omega +\delta
\end{split}
\end{equation}
This sampling grid dictates where to sample from the original input image, $I_{GI}$, for an output pixel in the output image, $I_{GW}$, which is the warped guiding image. After the calculation of $\rho$, the values are scaled back to $\left [ 0, h_{HR}-1 \right ]$ for $\delta (i,j,0)$ and $\left [ 0, w_{HR}-1 \right ]$ for $\delta (i,j,1)$ using the inverses of the previously given transforms. If we let $I(i,j,c)$ represents the pixel intensity value at the $(height=i,width=j,channel=c)$ location of the some image $I$, then the pixel intensity values at the output of the bilinear sampling module can be calculated using the following formula \cite{2015arXiv150602025J}: 
\begin{equation}
\label{eq:bilinear_sampling}
\begin{split}
I_{GW}(i,j,c)=\sum_{(a,b)\in H}I_{GI}(a,b,c)\\max\left \{ 0,1-\left | \rho (i,j,0)-a \right | \right \}\\max\left \{ 0,1-\left | \rho (i,j,1)-b \right | \right \}
\end{split}
\end{equation}
In equation \ref{eq:bilinear_sampling}, $H$ refers to the 4 closest pixel indices with respect to the coordinate given by $height=\rho (i,j,0)$ and $width=\rho (i,j,1)$. Wnet can be trained end-to-end with a loss function using gradient based methods due to the fact that $I_{GW}$ is sub-differentiable with respect to the parameters of the Wnet \cite{2015arXiv150602025J}.

\textbf{Generator (Gnet).} Gnet is the network that applies SR on the $I_{LR}$ while using the additional information provided by the warped guiding image $I_{GW}$. It consists of two smaller subnetworks, which are called SRnet and GFEnet. These two subnetworks represent two channels of information. SRnet takes only $I_{LR}$ as input, whereas GFEnet takes only $I_{GW}$ as input.

SRnet is the same baseline architecture used in \cite{2017arXiv170702921L}. The architecture is given in Figure \ref{fig:full_model_identity}. It consists of 16 residual blocks \cite{2015arXiv151203385H}, whose architecture can be found in the supplementary material. In our setting, scale parameter is set to $\alpha _{res}=1$, which is recommended in \cite{2017arXiv170702921L}. Throughout SRnet, spatial dimensions of $I_{LR}$ is preserved via zero padding. After the merging point of the global skip connection, upscaling blocks come, whose main responsibility is to gradually upscale the feature maps such that their spatial dimensions match with the spatial dimensions of $I_{GT}$. The upscaling is done via efficient sub-pixel convolutional layers \cite{2016arXiv160905158S}, that is in each upscaling block, $2\times$ upscaling is performed by cascading a convolutional layer and a pixel shuffler layer. These convolutional layers apply a $3\times3$ filter with stride 1 and they have 256 feature maps.

GFEnet, which is used as a feature extractor for the $I_{GW}$, consists of 3 downscaling blocks and 12 residual blocks.
As can be seen in Figure \ref{fig:full_model_identity}, each downscaling block is used to downscale the spatial dimensions of its input by 2. In each downscaling block, first, a convolutional layer with $3\times3$ kernel, 64 feature maps and stride 1 is applied, which is followed by a ReLU. Then another convolutional layer with $3\times3$ kernel, 64 feature maps and stride 2 is applied, which is again followed by ReLU. Downscaling of the $I_{GW}$ is done through series of stride 2 convolutions instead of max-pooling operation. The motivation is to let the model learn the downscaling procedure instead of fixing it \cite{2014arXiv1412.6806S}. After every 4th residual block, the current features that come from GFEnet and features that come from SRnet are fused. This feature fusion is done via concatenation along the depth axis. Since only convolutions with 64 output feature maps are used in both subnetworks, after the concatenation, a feature map of depth 128 is obtained. A convolution operation follows this concatenation before the signal resulting from the fusion operation enters to the next residual block of SRnet. After 12th residual block, which also means that after the 3rd feature fusion, GFEnet reaches to an end.

\textbf{Critic (Cnet).} The critic network is the same discriminator network that is used in DCGAN architecture ~\cite{2015arXiv151106434R} except we do not use batch normalization layers~\cite{2015arXiv150203167I}. The exact specifications of the architecture is given in the supplementary material.      

Outputs of Gnet, $I_{SR}$, form the generated samples, which should be criticized as fake images by the critic. The real images, which are samples from the real data distribution, are the same images that are used as the ground truth HR images for the LR input images. These should be criticized as real images by the critic.

\textbf{Identity Encoder (Inet).} We use a Siamese network \cite{Chopra:2005:LSM:1068507.1068961} for generating embedding vectors related with the identity of the person. We have selected VGG-16 network \cite{Simonyan14c} as the architecture of our siamese Inet, whose details can be found in the supplementary material. Given a SR face image $I_{SR}$ and its corresponding HR ground truth image $I_{GT}$, Inet is used to evaluate their similarity. This similarity information is then used to penalize $I_{SR}$ that has characteristics that differ from the characteristics of its corresponding $I_{GT}$. To learn the parameters $\theta_{Inet}$, we cast the problem as a binary classification problem, in which Inet tries to predict whether the two input images belong to the same person. This procedure can be guided by cross-entropy loss function, where the output is equal to
\begin{equation}
\label{eq:identity_binary_classification}
\begin{split}
y=sigmoid(w^{T}|Inet(x_{1};\theta_{Inet})-\\Inet(x_{2};\theta_{Inet})|+b)
\end{split}
\end{equation}
where $Inet(x;\theta_{Inet}), w\in\mathbb{R}^{4096}$ and $b\in\mathbb{R}$. Note that the parameters $w$ and $b$ are only used during pretraining of Inet. Moreover, during the optimization of Wnet, Gnet and Cnet, $\theta_{Inet}$ is frozen.

\subsection{Loss Functions}
\textbf{Content loss $L_{content}$.} Content loss is equal to L1 loss in our setting and can be calculated as:
\begin{equation}
\label{eq:content_loss}
\begin{split}
L_{content}=\frac{1}{3h_{HR}w_{HR}}\sum_{j=1}^{h_{HR}}\sum_{k=1}^{w_{HR}}\sum_{c=1}^{3}\\\left|I_{SR}(j,k,c)- I_{GT}(j,k,c)\right|
\end{split}
\end{equation}
$L_{content}$ ensures that contents of super-resolved image $I_{SR}$ match with those of $I_{GT}$. Although this loss is vital in keeping the connection between $I_{SR}$ and $I_{GT}$, and optimizes for high PSNR values, it results in $I_{SR}$ images that are formed by smooth regions and that lack high-frequency details \cite{8099502}.    

\textbf{Adversarial loss $L_{adv}$.} The aim of adversarial loss is to to make SR images look perceptually good and photo-realistic, making generated SR data distribution and real HR data distribution as close as possible to each other. With the help of the adversarial loss, the SR image will have fine details and the network will combat the smoothing effect caused by the content loss. 

The adversarial loss incurred by the WGAN-GP for the generator is equal to $-L_{fake}$. Note that $L_{fake}=D(I_{SR};\theta_{Cnet})$, where $\theta_{Cnet}$ represents the parameters of the critic and $D(I_{SR};\theta_{Cnet})$ represents the output of the critic for $I_{SR}$ image.

\textbf{Identity loss $L_{id}$.} Identity loss is calculated as the squared Euclidean norm of the distance between the embedding vectors of $I_{SR}$ and $I_{GT}$. Thus, 
\begin{equation}
\label{eq:identity_loss}
\begin{split}
L_{id}=\frac{\left \|Inet(I_{SR};\theta_{Inet})-Inet(I_{GT};\theta_{Inet})  \right \|^{2}}{4096}
\end{split}
\end{equation}

$L_{id}$ is used to penalize $I_{SR}$ that has characteristics that differ from the characteristics of its corresponding $I_{GT}$, thus increasing the perceptual quality of the SR image.

\textbf{Critic loss $L_{c}$.} We can calculate the loss incurred by the WGAN-GP for the critic as:
\begin{equation}
\label{eq:loss_critic}
\begin{split}
L_{c}=L_{fake}-L_{real}+\lambda _{gp}L_{gp}
\end{split}
\end{equation}
where $L_{real}=D(I_{GT};\theta_{Cnet})$ and $D(I_{GT};\theta_{Cnet})$ represents the output of the critic for $I_{GT}$ image. $\lambda _{gp}$ is the coefficient for the gradient penalty and $L_{gp}$, which is the gradient penalty term, is a function of $I_{SR}$, $I_{GT}$, $\theta_{Cnet}$ and $\epsilon \sim Uniform\left [ 0,1 \right ]$. Exact details are given in \cite{2017arXiv170400028G}. 

\textbf{Overall objective.} The overall objective function for the optimization of $\theta_{Wnet}$ and $\theta_{Gnet}$ can be written as:
\begin{equation}
\label{eq:overall_objective}
\begin{split}
L_{total}=L_{content}+\lambda_{adv}L_{adv}+\lambda_{id}L_{id}
\end{split}
\end{equation}
where $\lambda_{adv}$ and $\lambda_{id}$ are the weighting coefficients for $L_{adv}$ and $L_{id}$ respectively. The overall objective function for the optimization of $\theta_{Cnet}$ is directly equal to $L_{c}$. Optimization of these two objectives are done in an alternating fashion as also described in \cite{2017arXiv170400028G}. Note that during this training procedure $\theta_{Inet}$ is frozen.


\section{Experiments}
\subsection{Datasets}
\textbf{CelebA \cite{liu2015faceattributes}.} We used this dataset in developing our network and we moved to the dataset of \cite{Li_2018_ECCV} for comparing our method with the state-of-the-art. We use the aligned and cropped version of the CelebA dataset. We select the same train-validation-test partitioning used by the creators of the dataset. We have removed all of the identities that has a single image from the dataset, resulting in 162,734 training, 19,862 validation and 19,959 test images. It should be noted that the identities in each set are disjoint. To remove as much background as possible and to focus on the faces, we further crop the images to dimensions $168\times144$. For a given LR input image, the guiding image is sampled uniformly from the remaining HR images of the same person.

\textbf{Dataset of \cite{Li_2018_ECCV}.} This dataset is a subset of VggFace2 \cite{2017arXiv171008092C} and CASIA-WebFace \cite{2014arXiv1411.7923Y} datasets. All of the images are collected from the wild and therefore include different expressions, pose and illumination conditions. For each identity, pairs of HR guiding and ground truth images are available. All HR images have spatial dimensions $256\times 256$. Different from \cite{Li_2018_ECCV}, we randomly select 2,273 among 20,273 training images as validation images, which means that we are working with a smaller training set of size 18,000.

\subsection{Experiment Settings}
As a preprocessing step, we scale the input pixel intensity values from $\left [ 0,255 \right ]$ to $\left [ 0,1 \right ]$ and then subtract the mean of the training dataset. We also scale the range of the $I_{GT}$ to $\left [ -1,1 \right ]$. 

We always use Adam optimizer \cite{2014arXiv1412.6980K}. During training, whenever $\lambda_{adv}=0$, we use the suggested parameters $\beta _{1}=0.9$, $\beta _{2}=0.999$ and $\epsilon=10^{-8}$ \cite{2014arXiv1412.6980K}. Whenever $\lambda_{adv}\neq 0$, we use $\beta _{1}=0.5$ and $\beta _{2}=0.9$. During this adversarial training, we always apply $5$ critic updates per each generator update and we set $\lambda _{gp}=10$. 

\textbf{Training on CelebA.}
This dataset is used only during the development of the proposed method. The identity loss is not used during the training of GWAInet, thus $\lambda_{id}$ in Equation \ref{eq:overall_objective} is always set to 0. The training consists of three steps. We first pretrain the Wnet using L1 reconstruction loss between the warped guiding image and the ground truth image with learning rate $0.0001$ for 1.25 epochs with batch size 4. In the second step, we train the whole network by setting $\lambda_{adv}=0$ in Equation \ref{eq:overall_objective} with learning rate $0.0001$ and batch size 16 until the validation PSNR reaches its peak. Then we set $\lambda_{adv}=0.001$, and continue training for $4$ epochs using batch size 4 and then another $2$ epochs with learning rate $0.00005$.

\textbf{Training on the dataset of \cite{Li_2018_ECCV}.} We train the full model on this dataset. The training of GWAInet consists of two steps. We first train the network by setting $\lambda_{adv},\lambda_{id}=0$ in Equation \ref{eq:overall_objective} with learning rate $0.0001$ and batch size $48$ until the validation PSNR reaches its peak. Then we set $\lambda_{adv}=0.001$ and $\lambda_{id}=0.05$, and continue training for $8$ epochs using batch size 16. During the training of GWAInet, parameters of Inet is fixed. The Inet is pretrained on the same training set for 12 epochs with learning rate $0.0001$ and batch size 8.~\footnote{Our codes, models and results are publicly available on the project page: \url{https://github.com/berkdogan2/GWAInet}}

\begin{figure}
\centering
\setlength{\tabcolsep}{1pt}
\resizebox{\linewidth}{!}
{
\begin{tabular}{ccccc}
\includegraphics[width=.2\textwidth]{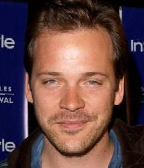}&
\includegraphics[width=.025\textwidth]{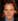}&
\includegraphics[width=.2\textwidth]{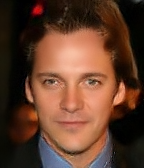}&
\includegraphics[width=.2\textwidth]{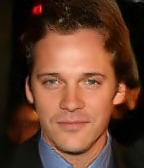}&
\includegraphics[width=.2\textwidth]{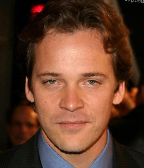}
\\
\includegraphics[width=.2\textwidth]{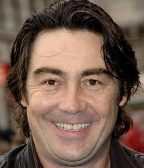}&
\includegraphics[width=.025\textwidth]{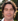}&
\includegraphics[width=.2\textwidth]{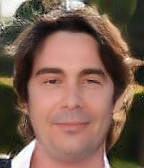}&
\includegraphics[width=.2\textwidth]{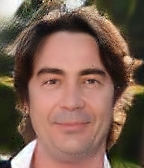}&
\includegraphics[width=.2\textwidth]{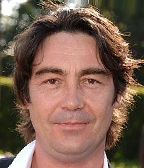}
\\
guiding face&
input &
w/o guiding face&
w/ guiding face&
ground truth
\\
&
&
BAnet result&
GWAnet result&
\end{tabular}
}
\vspace{-0.1cm}
\caption{CelebA results without (BAnet) and with (GWAnet) the use of the guiding face. [$8\times$ upscaling, LR input spatial dimensions $21\times18$]}
\label{fig:comp_guiding_celeba}
\end{figure}

\subsection{Results on CelebA}
After training on the CelebA dataset, we obtain the model GWAnet, which is the same model as the proposed full model GWAInet except that it does not include the identity loss in its optimization objective. Note that the identity loss is not related with the utilization of the guiding image because it is calculated between the super-resolved image and the ground truth image.

\begin{figure*}[!htbp]
\centering

\resizebox{0.8\linewidth}{!}
{
\begin{tabular}{cccccc}
    \includegraphics[width=.2\textwidth]{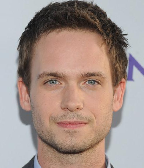}&
    \includegraphics[width=.2\textwidth]{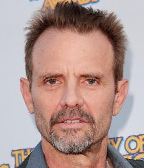}&
    \includegraphics[width=.025\textwidth]{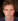}&
    \includegraphics[width=.2\textwidth]{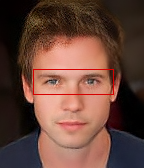}&
    \includegraphics[width=.2\textwidth]{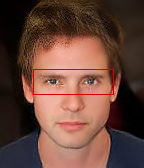}&
    \includegraphics[width=.2\textwidth]{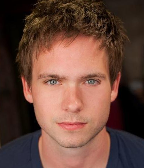}\\
\includegraphics[width=.2\textwidth]{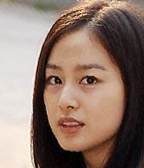}&
\includegraphics[width=.2\textwidth]{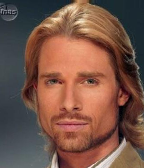}&
\includegraphics[width=.025\textwidth]{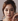}&
\includegraphics[width=.2\textwidth]{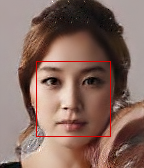}&
\includegraphics[width=.2\textwidth]{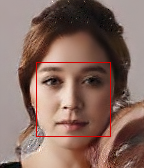}&
\includegraphics[width=.2\textwidth]{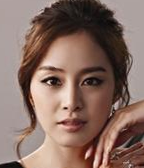}\\
    $I_{GI}$ &
    $I_{GI}$ &
    $I_{LR}$ &
    $I_{SR}$ &
    $I_{SR}$ &
    ground truth\\
    same identity &
    different identity&
    input&
    same identity &
    different identity&
    \\

\end{tabular}
}

\caption{Comparison of $I_{SR}$ face super resolved images for the cases when a guiding image with the same identity is used (GWAnet) and when a guiding image with a different identity is used (GWAnet-R) for CelebA dataset. [$8\times$ upscaling, LR input spatial dimensions $21\times18$]}
\label{fig:comp_random_celeba}

\end{figure*}

\textbf{Model without guiding image.}
To evaluate the importance of the guiding image and the subnetworks related with the guiding image, i.e. Wnet and GFEnet component of Gnet, we create a new model called BAnet, which only includes Cnet and SRnet component of Gnet. It is trained exactly in the same fashion as GWAnet. GWAnet, with the help of the guiding image, almost always improves the quality of the face image by adding some missing details about the facial features of the person over BAnet. GWAnet outperforms BAnet in generating perceptually good looking face images due to the fact that GWAnet provides better visual content quality by correcting the inaccurate facial details through the use of $I_{GI}$. In this context, visual content quality refers to the extent that the characteristics of the facial contents of $I_{SR}$ image match with those of $I_{GT}$ image. As can be seen in Figure \ref{fig:comp_guiding_celeba}, BAnet is fully capable of generating photo-realistic images as well as GWAnet but the point that sets GWAnet apart from BAnet is its ability to complete the missing facial details in $I_{SR}$ image by utilizing the guiding image $I_{GI}$. The improvements express themselves as location and shape improvements of facial features such as eyes, eyebrows, nose, mouth, hair and wrinkles.

\textbf{Guiding image with a different identity}
In order to evaluate the magnitude of the effect of the guiding image in generating $I_{SR}$, we have carried over an experiment, where for a given identity, we feed a randomly selected guiding image with a different identity. We denote the resulting model as GWAnet-R. The comparison of $I_{SR}$ images for GWAnet and GWAnet-R is shown in Figure~\ref{fig:comp_random_celeba}. In general, when the guiding image has a different identity, the resulting differences from the standard model are noticeable. For some cases, as exemplified by the second row in Figure~\ref{fig:comp_random_celeba}, the complete facial structure of the person changes. The mentioned differences mainly express themselves as location and shape differences of eyes and eyebrows as shown in the first row in Figure \ref{fig:comp_random_celeba}. There are also cases, where wrinkles appear or disappear according to the selected guiding image. The qualitative differences between GWAnet and GWAnet-R suggest that apart from being an additional information about high-resolution face images, the identity of the guiding image also plays an important role in generating high quality face images. 

\subsection{Comparison with state-of-the-art Methods}
\begin{figure*}[!htbp]
\begin{center}
\begin{subfigure}[t]{0.11\textwidth}
    \includegraphics[width=\textwidth]{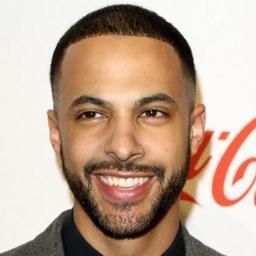}
    \includegraphics[width=\textwidth]{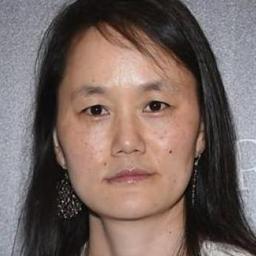}
    \includegraphics[width=\textwidth]{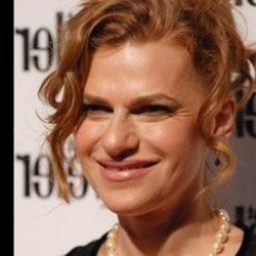}
    \includegraphics[width=\textwidth]{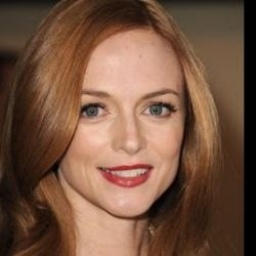}
    \includegraphics[width=\textwidth]{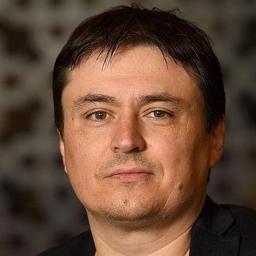}
    \caption*{Guiding face}
\end{subfigure}
\begin{subfigure}[t]{0.11\textwidth}
    \captionsetup{justification=centering}
    \includegraphics[width=\textwidth]{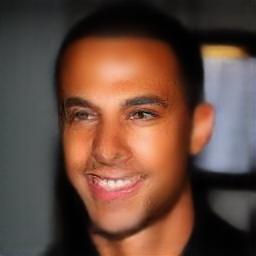}
    \includegraphics[width=\textwidth]{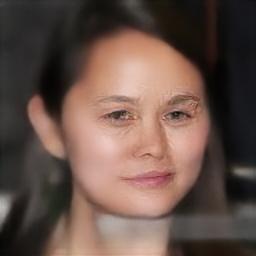}
    \includegraphics[width=\textwidth]{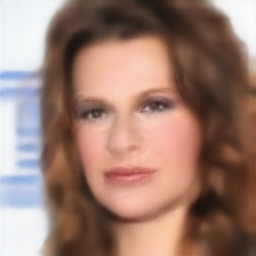}
    \includegraphics[width=\textwidth]{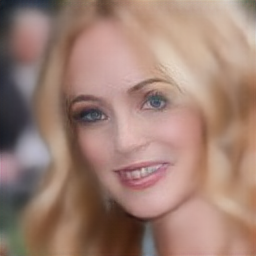}
    \includegraphics[width=\textwidth]{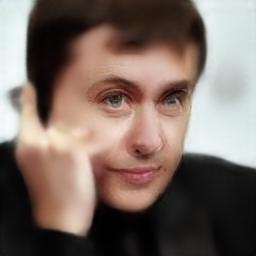}
    \caption*{GFRNet \cite{Li_2018_ECCV, 2018arXiv180404829L}}
\end{subfigure}
\begin{subfigure}[t]{0.11\textwidth}
   \includegraphics[width=\textwidth]{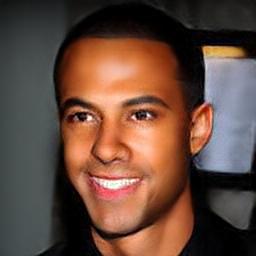}
    \includegraphics[width=\textwidth]{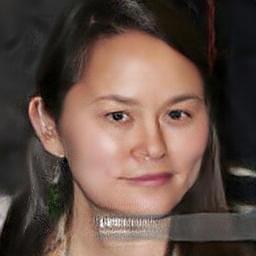}
    \includegraphics[width=\textwidth]{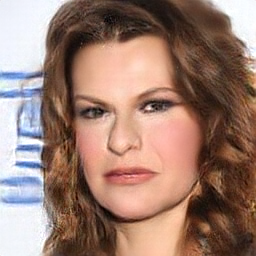}
    \includegraphics[width=\textwidth]{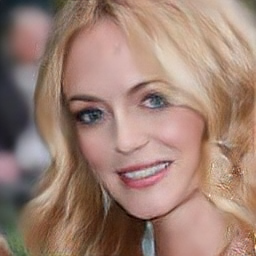}
    \includegraphics[width=\textwidth]{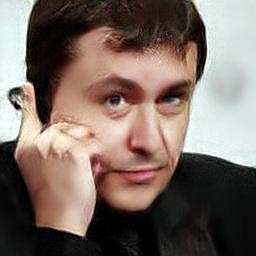}
    \caption*{Ours - Full}
\end{subfigure}
\begin{subfigure}[t]{0.11\textwidth}
   \includegraphics[width=\textwidth]{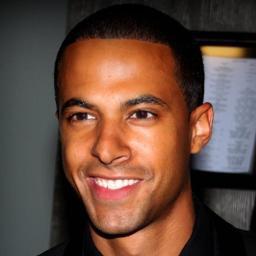}
    \includegraphics[width=\textwidth]{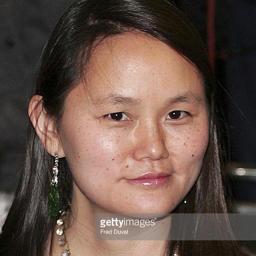}
    \includegraphics[width=\textwidth]{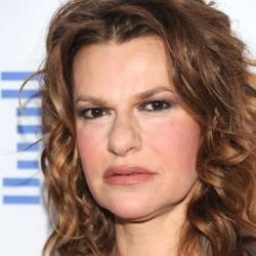}
    \includegraphics[width=\textwidth]{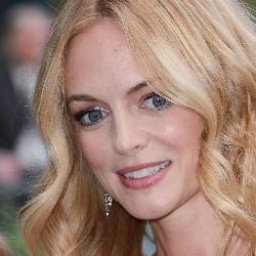}
    \includegraphics[width=\textwidth]{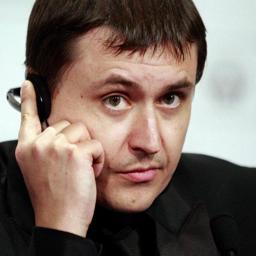}
    \caption*{Ground truth}
\end{subfigure}
\hspace{1em}
\begin{subfigure}[t]{0.11\textwidth}
    \includegraphics[width=\textwidth]{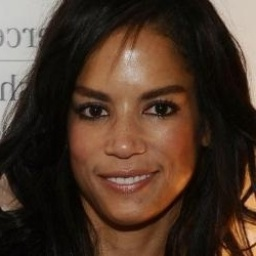}
    \includegraphics[width=\textwidth]{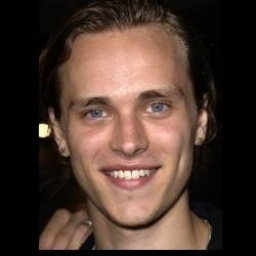}
    \includegraphics[width=\textwidth]{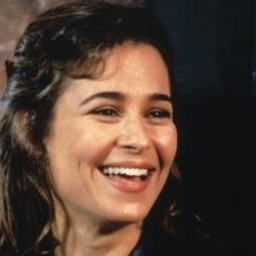}
    \includegraphics[width=\textwidth]{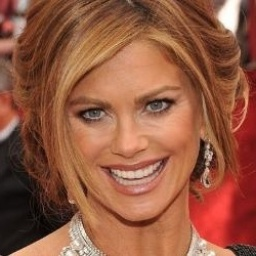}
    \includegraphics[width=\textwidth]{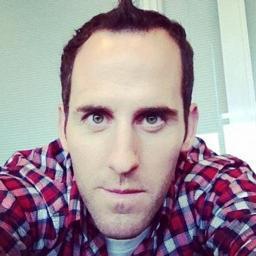}
    \caption*{Guiding face}
\end{subfigure}
\begin{subfigure}[t]{0.11\textwidth}
    \captionsetup{justification=centering}
    \includegraphics[width=\textwidth]{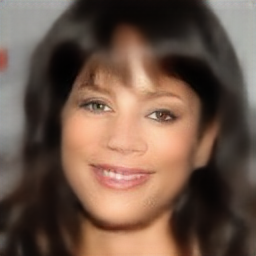}
    \includegraphics[width=\textwidth]{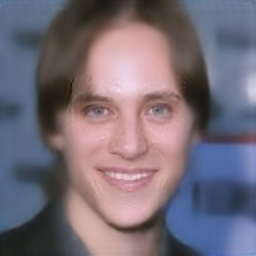}
    \includegraphics[width=\textwidth]{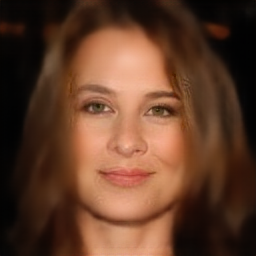}
    \includegraphics[width=\textwidth]{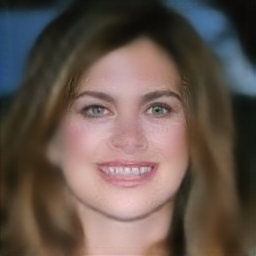}
    \includegraphics[width=\textwidth]{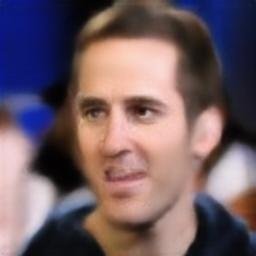}
    \caption*{GFRNet \cite{Li_2018_ECCV, 2018arXiv180404829L}}
\end{subfigure}
\begin{subfigure}[t]{0.11\textwidth}
    \includegraphics[width=\textwidth]{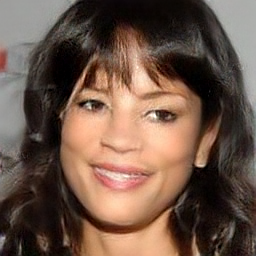}
    \includegraphics[width=\textwidth]{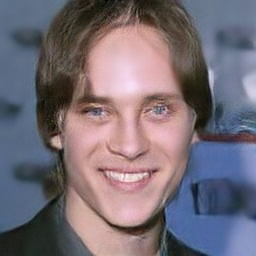}
    \includegraphics[width=\textwidth]{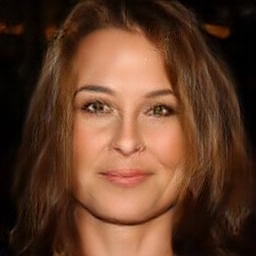}
    \includegraphics[width=\textwidth]{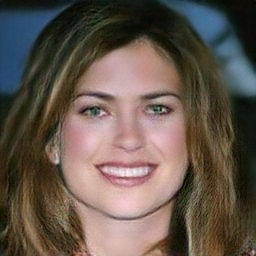}
    \includegraphics[width=\textwidth]{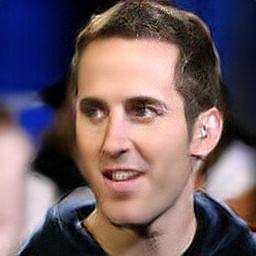}
    \caption*{Ours - Full}
\end{subfigure}
\begin{subfigure}[t]{0.11\textwidth}
    \includegraphics[width=\textwidth]{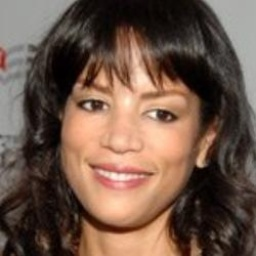}
    \includegraphics[width=\textwidth]{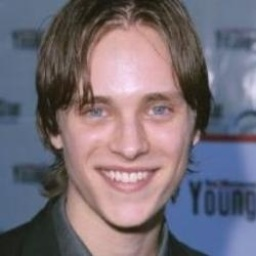}
    \includegraphics[width=\textwidth]{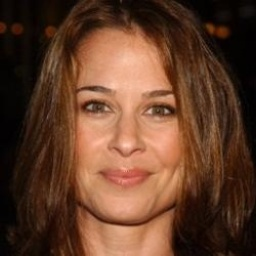}
    \includegraphics[width=\textwidth]{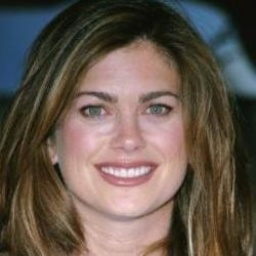}
    \includegraphics[width=\textwidth]{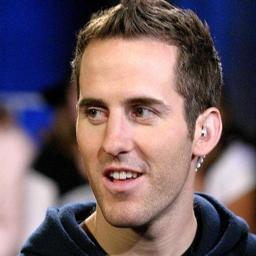}
    \caption*{Ground truth}
\end{subfigure}
\end{center}
\vspace{-0.25cm}
\caption{Comparison with state-of-the-art. Our full model GWAInet produces perceptually high quality images while retaining the facial features related with the identity. To obtain the results for GFRNet, their publicly available model is used \cite{Li_2018_ECCV, 2018arXiv180404829L}. [$8\times$ upscaling, LR input spatial dimensions $32\times32$]}
\label{fig:comp_main}
\end{figure*}

We compare our results quantitatively with the state-of-the-art face hallucination methods CBN \cite{10.1007/978-3-319-46454-1_37}, WaveletSR \cite{Huang2017WaveletSRNetAW}, TDAE \cite{Yu2017HallucinatingVL}, GFRNet \cite{Li_2018_ECCV} and super-resolution methods SRCNN \cite{10.1007/978-3-319-10593-2_13}, VDSR \cite{Kim2016AccurateIS}, SRGAN \cite{8099502}. For all those methods, we directly use the results reported in \cite{Li_2018_ECCV}. Moreover, we compare our results qualitatively with GFRNet \cite{Li_2018_ECCV}, which is the current state-of-the-art in face hallucination. Note that all of our experiments are performed for upscaling factor $8\times$. 

\textbf{Quantitative comparison.}
The quantitative results are shown in Table \ref{table:psnr_ssim_all}. As can be seen from Table \ref{table:psnr_ssim_all}, GWAInet outperforms the state-of-the-art in VggFace2 dataset by 1.47dB. It is the second best method in WebFace dataset and lags behind GFRNet \cite{Li_2018_ECCV} by 0.1dB. However, we should note that the training of GWAInet is not optimized for highest PSNR due to the adversarial loss and identity loss terms in the overall objective, which conflict with the objective of maximizing PSNR. PSNR is not well capable of capturing perceptual quality in an image \cite{8099502, Wang04imagequality, 1292216, 6146669}. Moreover, it is possible to get highest PSNR values by training GWAInet shorter or longer with very small decrease in perceptual quality. We did not follow such a path because the focal point of this paper is presenting the capability of GWAInet in producing perceptually high quality SR images.

\begin{table}[!htbp]
\centering
\begin{tabular}{||c||c|c|c|c|c|c||} 
\hline
Method & VggFace2 \cite{2017arXiv171008092C} & WebFace \cite{2014arXiv1411.7923Y} \\ [0.5ex] 
\hline\hline
SRCNN \cite{10.1007/978-3-319-10593-2_13} & 22.30 & 23.50  \\
VDSR \cite{Kim2016AccurateIS} & 22.50 & 23.65 \\ 
SRGAN \cite{8099502} & 23.01 & 24.49 \\ 
CBN \cite{10.1007/978-3-319-46454-1_37} & 21.84 & 23.10 \\ 
WaveletSR \cite{Huang2017WaveletSRNetAW} & 20.87 & 21.63 \\ 
TDAE \cite{Yu2017HallucinatingVL} & 20.19 & 20.24 \\ 
\hline
GFRNet \cite{Li_2018_ECCV} & \color{blue}24.10 & \color{red}27.21 \\ 
\hline
Ours (Full-GWAInet) & \color{red}25.57 & \color{blue}27.11 \\  
\hline
\end{tabular}

\caption[PSNR (dB) values for all models]{PSNR (dB) values for all models on the dataset of \cite{Li_2018_ECCV}. Upscaling factor is $8\times$ for all experiments. All results apart from the results of our models are taken from \cite{Li_2018_ECCV}. Red and blue markers indicate the first and second highest value, respectively.}
\label{table:psnr_ssim_all}
\end{table}

\textbf{Qualitative comparison.}
As can be seen from Figure~\ref{fig:comp_main}, our method GWAInet produces better looking and sharper face images than the state-of-the-art. GFRNet \cite{Li_2018_ECCV} only sharpens a small area in the face region, whereas our method GWAInet introduces high frequency details for all parts of the image, including the hair. Moreover, GFRNet \cite{Li_2018_ECCV} generally completely hallucinates the face of the person such that the super-resolved face does not look like the same identity. Our method, on the other hand, is completely faithful to the identity of the person while super-resolving the face image. Furthermore, GFRNet \cite{Li_2018_ECCV} most of the time outputs a super-resolution image that is blurry and that contains artifacts, whereas our method GWAInet produces sharp, visually appealing and photo-realistic results.

\section{Conclusion}
We proposed a novel solution, namely GWAInet, for the task of face image super-resolution.
Our GWAInet utilizes the additional information provided by a high-resolution guiding image of the same person. Our network does not use facial landmarks during training and is capable to produce fine details for the whole face region in a uniform manner. Moreover, in the training, the employed identity loss further helps in preserving the identity related features by minimizing the distance between the embedding vectors of the super-resolved and HR ground truth images. GWAInet produces photo-realistic images in upscaling factor $8\times$ and outperforms state-of-the-art in PSNR terms and also for perceptual quality of super-resolved images.

\section*{Acknowledgments }
This work was partly supported by ETH Zurich General Fund (OK), Huawei, and a GPU grant from Nvidia.

{\small
\bibliographystyle{ieee}
\bibliography{egbib}
}

\newpage
\appendix
\section{Supplementary Material}
This supplementary material provides the details on the network architectures used in our proposed solution from the main paper.

\subsection{Our Network Architectures}

Table~\ref{table:architectural_details} provides the descriptions of the Warper (Wnet), Critic (Cnet), and Identity Encoder (Inet) subnetworks as employed in our proposed GWAInet (see Fig.~\ref{fig:full_model_identity}).

\begin{table}[!htbp]
\centering
\begin{tabular}{||c|c|c||} 
\hline
Wnet & Cnet & Inet \\ [0.5ex] 
\hline\hline
k3n64s1*& k5n64s2** & k3n64s1* \\
k3n64s2*& k5n128s2** & k3n64s1* \\
k3n64s1*& k5n256s2** & max\_pool(k2s2) \\
k3n64s2*& k5n512s2** & k3n128s1* \\
k3n64s1*& fc(1) & k3n128s1* \\
k3n64s2*& & max\_pool(k2s2) \\
k3n64s1& & k3n256s1* \\
skip start& & k3n256s1* \\
8x ResBlock & & max\_pool(k2s2) \\
k3n64s1& & k3n512s1* \\
skip end & & k3n512s1* \\
pixel shuffler 2x& & k3n512s1* \\
pixel shuffler 2x& & max\_pool(k2s2) \\
pixel shuffler 2x& & k3n512s1* \\
k3n2s1& & k3n512s1* \\
& & k3n512s1* \\
& & max\_pool(k2s2) \\
& & fc(4096)* \\
& & fc(4096)*\\
& & fc(4096) \\
\hline
\end{tabular}
\caption{Architectures of Wnet, Cnet and Inet. Note that * and ** symbols refer to ReLU and LeakyReLU ($\alpha=0.2$) layers, respectively. k3n64s1 represents a convolution operation with kernel size 3x3, 64 feature maps and stride 1.}
\label{table:architectural_details}
\end{table}

\subsection{Residual Block}

\begin{figure}[htbp!]
\centering
\includegraphics[width=0.5\linewidth]{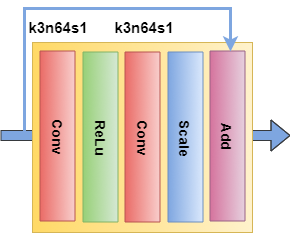}
\vspace{-0.2cm}
\caption{Residual block as introduced in \cite{2017arXiv170702921L}}
\label{fig:res_block}
\vspace{-0.2cm}
\end{figure}

The structure of the residual block is shown in Figure~\ref{fig:res_block}.

\end{document}